# Linguistic traces of stochastic empathy in language models

**Bennett Kleinberg[1,2*], Jari Zegers[1], Jonas Festor[1], Stefana Vida[1], Julian Präsent[3], Riccardo Loconte[1,4], Sanne Peereboom[1]**

**[1] Tilburg University, Department of Methodology and Statistics, The Netherlands**
**[2] University College London, Department of Security and Crime Science, UK**
**[3] University of Amsterdam, Department of Psychology, The Netherlands**
**[4] IMT School of Advanced Studies, Lucca, Italy**

**Abstract**

Differentiating generated and human-written content is increasingly difficult. We examine how an incentive to convey humanness and task characteristics shape this human vs AI race across five studies. In Study 1-2 (*n*=530 and *n*=610) humans and a large language model (LLM) wrote relationship advice or relationship descriptions, either with or without instructions to sound human. New participants (*n*=428 and *n*=408) judged each text's source. Instructions to sound human were only effective for the LLM, reducing the human advantage. Study 3 (*n*=360 and *n*=350) showed that these effects persist when writers were instructed to avoid sounding like an LLM. Study 4 (*n*=219) tested empathy as mechanism of humanness and concluded that LLMs can produce empathy without humanness and humanness without empathy. Finally, computational text analysis (Study 5) indicated that LLMs become more human-like by applying an implicit representation of humanness to mimic stochastic empathy.

## INTRODUCTION

Large language models (LLMs) have been among the landmark technical achievements of the past decade [1–3]. With LLMs' increasing ability to generate text comes the challenge for consumers of digital content to tell whether a text was written by a human or generated by an AI model. This paper argues that previous comparisons between LLMs and humans in producing convincing prose may have been unjust to human language ability. We adopt experimental research methods from psychology to investigate whether the need to use empathy and an incentive to appear human have overestimated LLM's ability to impersonate human writing in previous work.

Digital content is increasingly produced by or with the help of large language models, and recent evidence indicates that LLM-generated content is highly effective in influencing opinion – both in an encouraging form shown to dampen conspiratorial thinking in humans [4] and more concerningly, in the form of persuasive messaging on political issues [5]. Consequently, humans as consumers of an ever-growing amount of digital content increasingly aplenty with AI-generated pieces are facing the task of distinguishing human from machine-generated content.

There is evidence that it is increasingly difficult for humans to tell human from AI-generated content for images [6], voice [7] and text [8,9]. In the domain of textual data, recent work compared human-written online self-representations (dating profiles, professional bio sketches, hospitality profiles) to those generated by a language model [9]. Independent human assessors then evaluated all texts. The findings suggested that humans often use misguided cues to tell AI from humans (e.g., incorrectly associating rare words and grammatical issues with LLM-generated content) and that exploiting these flawed heuristics can be used to source AI content that is considered more human than that written by humans. The current paper builds on that study and extends that paper in two aspects.

First, human authors wrote the texts without any incentive to appear human in the face of AI technology capable of imitating human writing with the skill and quality that current LLMs show. The study was conducted at the early stage of generative text models (i.e., using GPT-2 and GPT-3) before these technologies reached mainstream attention with the launch of ChatGPT in December 2022. While humans do have incentives to appear genuine relative to other humans who misrepresent details about themselves (e.g., in dating profiles [10]), the possibility of an automated system (e.g., an LLM) mimicking human writing and masquerading as humans effectively at scale is rather novel. Put differently, an incentive to appear human was arguably less relevant several years ago before a widespread adoption of LLMs, but this may be about to change with more usage and knowledge about language models.

Second, the context of the writing task was constrained to self-presentations which carry their own communicative goals. Such self-representation tasks might limit the degree to which humanness can "shine through" because writers are focused on impression management in addition to expressing themselves (e.g., presenting oneself favourably, eliciting contacts for dating) - a self-serving objective that is not disentangled from other objectives of self-representation (e.g., the need to navigate between acceptable and intolerable misrepresentations in dating profiles [10]). Self-presentations are human-oriented but also serve other objectives. Our goal was to find a different task that is open-ended to allow humans to resort to the full breadth of human language ability without these other objectives interfering. One of the qualities that such a task should allow to transpire is empathy since it partially requires nuanced human understanding of another person but also contains components that are deemed achievable for LLMs [11].

Empathy is a multifaceted construct and some have argued that using empathy as an umbrella term for its many meanings causes unnecessary unclarity [12]. According to [12,13], important to distinguish are three aspects of empathy: cognitive (i.e., understanding and inferring emotions of others) and emotional empathy (i.e., experiencing the feelings of another person or group even if they are not present) as well as motivational empathy, which some [12] argue should be replaced by compassion for others (i.e., wishing another person well). Others have used an overarching differentiation between "self-focused" empathy and "other-focused" compassion [14]. Accordingly, empathy involves making another person's feelings the focus of one's own attention, while compassion focuses exclusively on the other person's mental and physical well-being. Discussions of "empathic AI" seem to agree that since emotional empathy requires experiencing feelings of others, current LLMs would not be expected to genuinely possess emotional empathy [11,15]. Compassion and concern for others pose an additional challenge due to the motivational component involved (i.e., a willingness to invest time or other resources in improving the other's well-being). Cognitive empathy, however, would not require experiencing an emotion but rests on identifying and acknowledging another person's emotions. Recent work where human participants were receiving emotional support from an LLM (Bing Chat) suggest that humans indeed report feeling heard as long as they were led to believe the response stemmed from another human [16] (see also [17]). Cognitive empathy might thus be the simplest empathy challenge for LLMs but even emotional empathy could effectively be faked [11]. The goal of the current work is not to contribute to the definitional debates in empathy research [12] but, instead, to use empathy as a playground on which humans and LLMs meet to convey humanness through writing. In this paper, we use relationship advice as the domain of interest that meets that requirement [18]. We treat relationship advice not as a manipulation of empathy but as a task that increases the relevance of empathic reasoning. Humans should be able to convey humanness when providing relationship advice for their ability to experience the full breadth of empathy, while an LLM should only be able to resort to cognitive empathy. To put this to the test, we use experimentation approaches on humans and LLMs borrowed from recent trends in the study of machine behaviour.

Approaches to studying LLMs have recently adopted research methods and designs from the social and psychological sciences. Some argue that AI models – including LLMs – ought to be studied analogous to how animal and human behaviour is investigated through behavioural patterns in an artificial, social environment [19]. Borrowed from early behaviourist ideas, the machine behaviour school is careful not to attribute inner processes to how AI models, such as LLMs, work. A more direct reference to the field of psychology is made elsewhere [20] with the argument that differences in observable behaviour (e.g., generated text) attributable to manipulation in input (e.g., changing prompt instructions) can provide insights about underlying processes or representations (e.g., bias). Consequently, the experiment is the most informative research design for studying the behaviour and processes of LLMs. Using that approach led to the discovery of racial prejudices in LLMs [21], proneness to cognitive biases [22], illusory truth effects in LLMs [23] and early efforts on testing a theory of mind in AI models [24,25]. For the current paper, we adopt the experimental method and apply it to both human participants and LLM queries.

While there is thus evidence that human assessors struggle to distinguish human-written content from LLM-generated content, two important moderators of that effect have yet to be examined. First, it is still being determined whether humans can convey their 'humanness' when they are aware of an AI system trying to mimic human writing. Second, the characteristics of the writing task may affect the degree to which an LLM can convey humanness. Previous work has used tasks that are relatively low on required empathy and, hence, potentially more manageable for an LLM. We experimentally disentangle the effects of human task awareness (Study 1) and required empathy (Study 2). The robustness of the results from Study 1 and 2 was tested in a third experiment (Study 3) where we examine the effect of the phrasing of task instructions. Study 4 then tested whether empathy is a mechanism through which humanness is conveyed. Using computational text analysis, we then shed light on the strategies to appear human (Study 5).

## STUDY 1

We aimed to isolate the effect of human task awareness on a writing task that required a nuanced understanding of human relationships. Participants were tasked with writing relationship advice whilst either being made aware (adversarial condition) or not (naïve condition) of an AI system performing the same task to appear as human as possible. This experiment resulted in a 2 (Source: human vs LLM) by 2 (Condition: adversarial vs naive) between-subjects design with the source evaluation of a new sample of humans as the dependent variable. The sample size for this experiment was based on an *a priori* power analysis for a small effect size ($f$=0.10), an alpha threshold of 0.01, and a power of 0.80, resulting in a required sample size of 1172 (i.e., 586 human-written texts, 586 LLM-generated texts).

*Ethics and transparency statement*

The ethics review board of [ANONYMISED FOR PEER REVIEW] approved all experiments reported in this paper before data collection. All participants provided informed consent at the beginning of each experiment. All data, materials and code to reproduce the LLM data collection are publicly available at:
https://osf.io/rcz6s/?view_only=82e087f3c0a64b0aaf6039094cfc5c9c.

Method

*Human-written vs LLM-generated relationship advice*

We recruited a sample of $n$=587 participants from Prolific (50.77% female, $M_{age}$=30.15 years, $SD_{age}$=10.80). Each participant was randomly assigned to either the adversarial ($n$=315) or the naive condition ($n$=272). We excluded 57 participants who failed the manipulation check asking about their condition assignment.

All participants provided informed consent, received task instructions, and performed the writing task. The writing task provided them with general details about a fictitious relationship, consisting of the relationship type (heterosexual vs. homosexual), length of the relationship (3 months vs. 20 years), and reason for conflict (cheating vs. moving to another continent), resulting in eight stimulus sets. Participants were randomly allocated to one stimulus set and then typed their relationship advice with a minimum length of 500 characters.

Upon completion of the writing task, participants answered a manipulation check question ("What was your task?") selecting one out of four answer options ("Write relationship advice," "Simple creative writing task," "Write relationship advice while appearing as human as possible," and "Write a LinkedIn bio") and indicated – on a 7-point Likert scale (1 – completely disagree, 7 – completely agree) – how difficult they found the task and how motivated they were to perform well. All participants were debriefed and finally redirected to Prolific. Each participant was remunerated with GBP 1.25.

We collected corresponding relationship advice from the GPT-4 language model (model name: gpt-4-0125-preview). Via the API, the LLM was prompted with instructions identical to those of the human sample. Each request was sent with a randomly chosen sampling temperature (i.e., the degree to which the model selects subsequent tokens with a lower probability—higher temperatures indicate higher randomness of the model's text completions).[1]

*Adversarial vs naive motivation*

In both author modalities (human vs LLM), the naive group received basic instructions on writing the relationship advice. Specifically, that group was instructed to "[...] *write a brief piece of advice to a friend who is having relationship problems. Basic details of the scenario will be provided. The advice should be about what the person should do next in an upcoming confrontation, which may lead to a breakup."*

The adversarial group received task instructions that made them aware that humans would read each text and determine its source. It was stressed that their goal was to appear human and not AI-generated. The instructions read as follows: "*Important: write advice that comes across as human-written, and not AI-generated. Participants in a future study will read your advice and indicate whether they think it is AI-generated. Your goal is to write a text that human participants will perceive as human-written*" (see Supplementary Material SM 1 for verbatim examples).

*Source evaluation*

To evaluate the final sample of 1060 texts produced by humans (*n*=530) and LLM (*n*=530) under the abovementioned conditions, we recruited a new sample of participants from Prolific (*n*=428, 47.43% female, $M_{age}$=31.27, $SD_{age}$=9.62). Each participant read a random selection of ten texts and indicated their judgment about the source on the following 5-point scale [9]: "1=definitely AI-generated", "2=likely AI-generated", "3=not sure", "4=likely human-written", "5=definitely human-written". Participants were remunerated with GBP 2.00. Each statement was evaluated on average 3.44 times (*SD*=0.71). We used the mean of the ratings for each text to arrive at the dependent variable source judgment for each text (i.e., collapsing judgments by multiple judges to one average judgment).

*Analysis plan*

For the main analysis, we report the findings as an ANOVA on the text-level mean-aggregated judgments. As a robustness check, we also report the multilevel findings on the non-aggregated data in the supplementary material (SM 14) where we include text and human rater as random intercepts. These analyses corroborate the ones reported here.

---

[1] The decision to randomly sample the temperature was motivated by the absence of any clear guideline or evidence about the "correct" or recommended temperature to choose (see SM 13).

Further, for contrasts comparing two groups, we report Frequentist null hypothesis significance tests with Cohen's *d* effect sizes and 99% confidence intervals, as well as Bayes Factors (BF) derived from Bayesian t-tests [26] using the *BayesFactor* R package with default priors [27]. The BF quantifies relative evidence for one hypothesis over another hypothesis and expresses how much more likely the data are given one hypothesis compared to the other. We report $BF_{10}$ and $BF_{01}$, which express evidence in favour of the alternative hypothesis (i.e., that there is a difference between two groups) over the null hypothesis, and evidence in favour of the null over the alternative hypothesis, respectively. For $BF=1.00$, both hypotheses are equally supported. We adhere to Bayes Factor interpretation guidelines that identify anecdotal evidence as $1<BF<3$, moderate evidence as $3<BF<10$, strong evidence as $10<BF<30$ and extreme evidence as $BF>100$ [28].

## Results

### *Preliminary analysis*

Participants perceived the task to be of low-to-moderate difficulty (adversarial condition: $M$=3.50, $SD$=1.79; naive condition: $M$=3.45, $SD$=1.76), with a *t*-test indicating no significant difference between the two conditions, $d$=0.03 [99%CI: -0.20; 0.25], $BF_{10}$=0.10, $BF_{01}$=9.90. Similarly, there was no significant difference in motivation to perform well (adversarial condition: $M$=5.81, $SD$=1.16; naive condition: $M$=5.73, $SD$=1.26), $d$=0.07 [-0.16; 0.29], $BF_{10}$=0.13, $BF_{01}$=7.70, with participants indicating a high motivation overall.

### *Human judgments*

The 2 (Source) by 2 (Condition) between-subjects ANOVA revealed a significant main effect of Source, $F(1,1056)=203.46$, $p<.001$, $\eta^2=0.16$ [0.12, 1.00]. That effect showed that texts written by humans ($M$=3.46, $SD$=0.89) were rated as more human than those generated by an LLM ($M$=2.75, $SD$=0.85), $d$=0.82 [0.65; 0.98], $BF_{10}>100$, $BF_{01}<0.01$, regardless of the task instructions. A significant main effect of Condition, $F(1, 1056)=99.31$, $p<.001$, $\eta^2=0.09$ [0.05; 1.00], indicated that texts written under the adversarial instructions ($M$=3.34, $SD$=0.82) were perceived as more human than texts written under the naive instructions ($M$=2.85, $SD$=0.99), $d$=0.55 [0.39; 0.71], $BF_{10}>100$, $BF_{01}<0.01$. That effect was subsumed under the significant interaction (Figure 1) between Source and Condition, $F(1, 1056)=65.69$, $\eta^2=0.06$ [0.03; 1.00], revealing that the adversarial instructions had no significant effect for humans, $d$=0.10 [-0.12; 0.33], $BF_{10}$=0.19, $BF_{01}$=5.14, but a substantial significant effect LLMs, $d$=1.24 [1.00; 1.49], $BF_{10}>100$, $BF_{01}<0.01$.

Figure 1. *Human judgments for relationship advice by Source (LLM vs human) and Condition (naïve vs adversarial) in Study 1*.

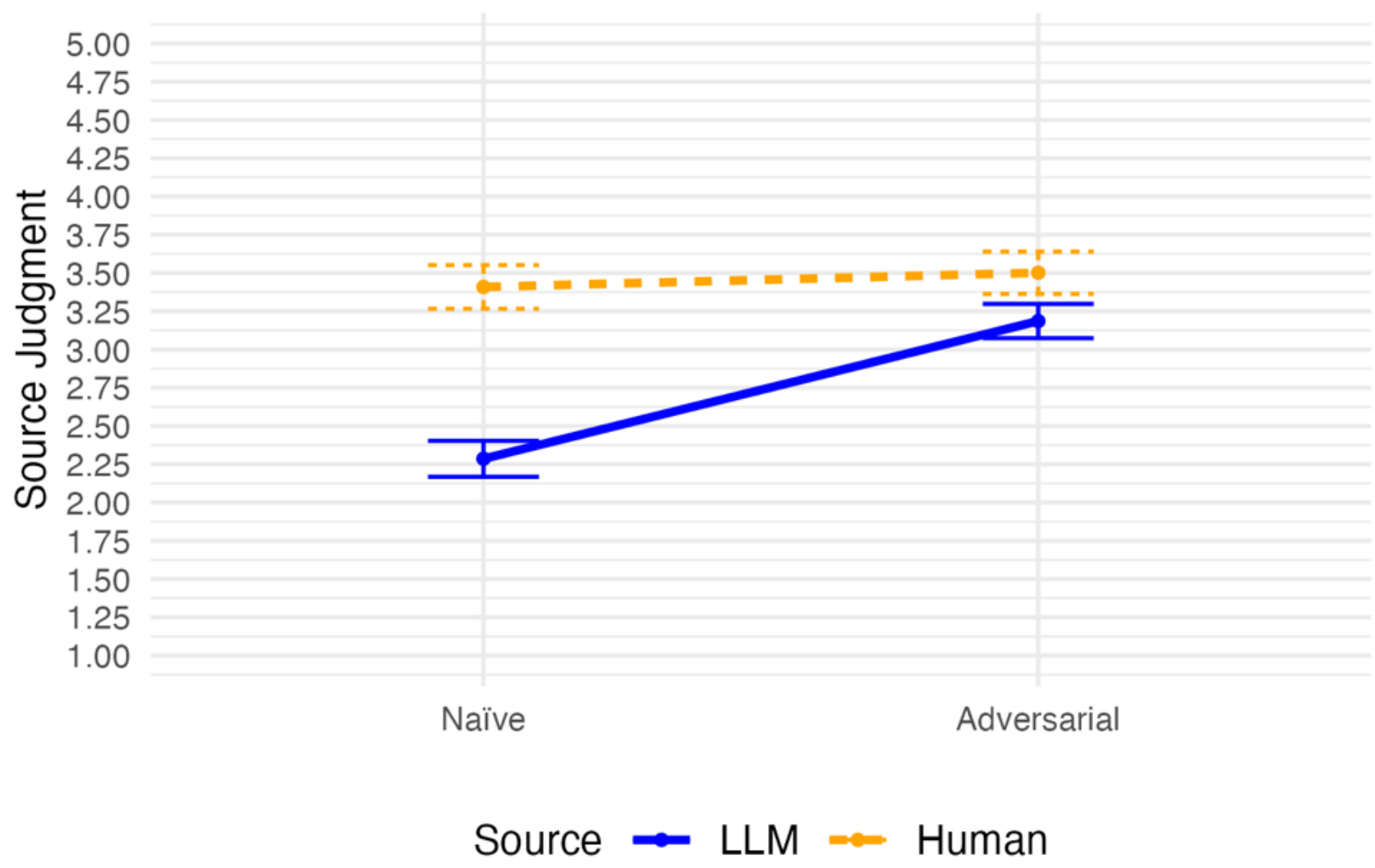

Note. The bars around the means represent the 99% confidence intervals. Source judgment values ranged from 1=definitely AI-generated to 5=definitely human-written. See SM2 for results in table format.

*Diagnostic value of human judgments*

We used receiver operating characteristics to assess how well human judgments can differentiate between human-written and LLM-generated texts and compared the areas under the curve (AUCs). Receiver operating characteristics analysis assesses how well a response variable (here: human judgments) detects a categorical signal variable (here: LLM-generated vs human-written). The AUC is obtained by plotting the sensitivity (i.e., true positive rate) against 1-specificity (i.e., true negative rate) across all response variable values as classification threshold values and measures how much the area under that curve deviates from a non-diagnostic diagonal line. The decisive advantage of that approach is that is threshold-independent and does not require a specific threshold value as is the case in accuracy or other classification metrics. The AUC ranges from 0 to 1, with values closer to 1 indicating higher diagnostic value and a value of 0.50 indicating no diagnostic value of the response in detecting the signal.

The diagnosticity of human judgments in telling the source of the texts written under naive instructions was high, AUC=0.83 [0.78; 0.88]. However, under the adversarial condition, this dropped markedly, AUC=0.62 [0.56; 0.68]. The DeLong's test [29,30] for comparing two AUCs indicated a significant difference, $D(952.92)=7.05$, $p<.001$.

Discussion Study 1

Human-written pieces of relationship advice were perceived as more human than those written by an LLM. Notably, only the LLM responded to the adversarial instructions: when instructed to be as human as possible, the LLM could drastically increase the humanness perception (+39.3%) while humans could not (+2.6%). Although we presumed that the relationship advice task required more empathy than a descriptive writing task, we lack decisive evidence for that claim. To address this, we experimentally manipulated the writing task in Study 2.

## STUDY 2

In the second experiment, we added an experimental manipulation of the characteristics of the writing task. We contrasted the relationship advice (identical to Study 1) with a control task. The advice task afforded greater opportunities for empathic reasoning (e.g., perspective-taking, sensitivity to others' needs) than merely describing a relationship. That additional factor resulted in a 2 (Source: human vs LLM) by 2 (Condition: adversarial vs naive) by 2 (Task: advice vs description) between-subjects design. The sample size was based on a power analysis for $f$=0.15, $\alpha$=0.01, and power=0.80, resulting in a total sample size of 523 for the human sample. We obtained the same number of texts from the LLM.

### Method

The experimental conditions followed the same procedure as in Study 1 and we recruited participants (who were not participating in Study 1) from Prolific ($n$=610, 50.16% female, $M_{age}$=34.92, $SD_{age}$=12.53, remuneration: GBP 1.25) who were assigned to either of the four experimental conditions: relationship advice and control condition ($n$=159), relationship advice and adversarial condition ($n$=160), relationship description and control ($n$=149) or relationship description and adversarial ($n$=142). Those that failed the manipulation check were excluded ($n$=78).

#### *Relationship advice vs relationship descriptions*

Participants in the relationship advice condition followed the same procedure as those in Study 1. In the relationship description condition, participants were told to "*[...] write a brief description of a friend's relationship who is having relationship problems. Basic details of the scenario will be provided. The description should be based on those details and written in full sentences*". Identical to Study 1, humans and the GPT4 model received the same instructions in each condition.

#### *Additional control questions*

We further included the following control variables in the human writing task: the participants' self-rated ability to express themselves well in the text they wrote and whether the task required much empathy (both on a 7-point scale from "completely disagree" to "completely agree").[2]

#### *Source evaluation*

A sample of 406 participants (52.96% female, $M_{age}$=29.35, $SD_{age}$=8.26, remuneration: GBP 2.00) recruited from Prolific evaluated the 1064 texts (half of which were human, half LLM-generated) using the 5-point scale from Study 1 ("1=definitely AI-generated" to "5=definitely human-written"). On average, 3.24 humans assessed each statement ($SD=0.65$).

[2] We also measured participant-level variables in the human evaluation task and obtained self-reported regularity of usage of LLMs, experience with creating LLM text, experience with consuming LLM text, their ability to spot LLM written content, and general knowledge about LLMs. These were not the focus of the current work but are available in the public dataset.

Results

*Preliminary analysis*

Participants indicated a high motivation to perform well (SM3), a high ability to express themselves and a moderately high degree of empathy required. There were no significant differences between conditions in self-rated expression ability ($p=.437$), $BF_{10}=0.13$, $BF_{01}=7.73$, or motivation ($p=.147$), $BF_{10}=0.27$, $BF_{01}=3.71$. For the perceived level of empathy required, there was a significant main effect of Task ($F(1, 528)=31.46$, $p<.001$, $\eta^2=0.06$ [0.02; 1.00]), with the advice task ($M=5.61$, $SD=1.27$) being perceived as requiring significantly more empathy than the description task ($M=4.94$, $SD=1.50$), $d=0.49$ [0.26; 0.72], $BF_{10}>100$, $BF_{01}<0.01$. Similarly, perceived difficulty differed by Task, $F(1, 528)=23.65$, $p=.003$, $\eta^2=0.02$ [0.00; 1.00], so that the relationship description task ($M=3.85$, $SD=1.69$) was rated as significantly more difficult than the advice task ($M=3.42$, $SD=1.67$), $d=0.26$ [0.03; 0.48], $BF_{10}=6.29$, $BF_{01}=0.16$.

*Human judgments*

The 2 (Source) by 2 (Condition) by 2 (Task) ANOVA showed a significant three-way interaction between these factors, $F(1, 1056)=29.93$, $p<.001$, $\eta^2=0.03$ [0.01; 1.00], under which the other effects were subsumed. We unpack that three-way interaction as the main result by Task.

For the relationship description task, the Source by Condition ANOVA indicated no significant interaction. There was a significant main effect of Source, $F(1, 450)=24.34$, $p<.001$, $\eta^2=0.05$ [0.01; 1.00]. Relationship descriptions written by humans ($M=3.29$, $SD=0.82$) were perceived as significantly more human than LLM-generated ones ($M=2.93$, $SD=0.75$), $d=0.46$ [0.22; 0.71], $BF_{10}>100$, $BF_{01}<0.01$ (Figure 2).

For the relationship advice task, the analysis revealed a significant main effect of Source, $F(1, 606)=179.40$, $p<.001$, $\eta^2=0.23$ [0.16; 1.00], indicating that texts originating from humans ($M=3.59$, $SD=0.87$) were rated as significantly more human than LLM-generated texts ($M=2.70$, $SD=0.89$), $d=1.01$ [0.70; 1.23], $BF_{10}>100$, $BF_{01}<0.01$. For condition, the significant main effect, $F(1, 606)=43.01$, $p<.001$, $\eta^2=0.07$ [0.03; 1.00], suggested that relationship advice written under the adversarial instruction ($M=3.36$, $SD=0.87$) was rated as significantly more human than those written under the naive instructions ($M=2.93$, $SD=1.04$), $d=0.45$ [0.24; 0.66], $BF_{10}>100$, $BF_{01}<0.01$.

The critical interaction replicated the key effect from Study 1, $F(1, 606)=53.89$, $p<.001$, $\eta^2=0.08$ [0.04; 1.00]: for the relationship advice task, the adversarial instructions did not significantly affect humans, $d=0.06$ [-0.35; 0.24], $BF_{10}=0.14$, $BF_{01}=6.97$, but had a substantial significant effect on LLMs, $d=1.21$ [0.89; 1.53], $BF_{10}>100$, $BF_{01}<0.01$.

*Diagnostic value of human judgments*

The AUC analysis replicated the findings from Study 1. For the advice task, the diagnosticity under the naive instructions was high, AUC=0.88 [0.83; 0.93], but dropped significantly under the adversarial instructions, AUC=0.63 [0.55; 0.72], $D(505.99)=6.57$, $p<.001$. In comparison, the AUC for the relationship description task was considerably lower (naive: AUC=0.60 [0.50; 0.70]; adversarial: AUC=0.65 [0.56; 0.74]) and did not differ significantly between conditions, $p=.354$.

Figure 2. *Human judgments in Study 2 for relationship advice (left) and relationship descriptions (right) by Source (LLM vs human) and Condition (naïve vs adversarial).*

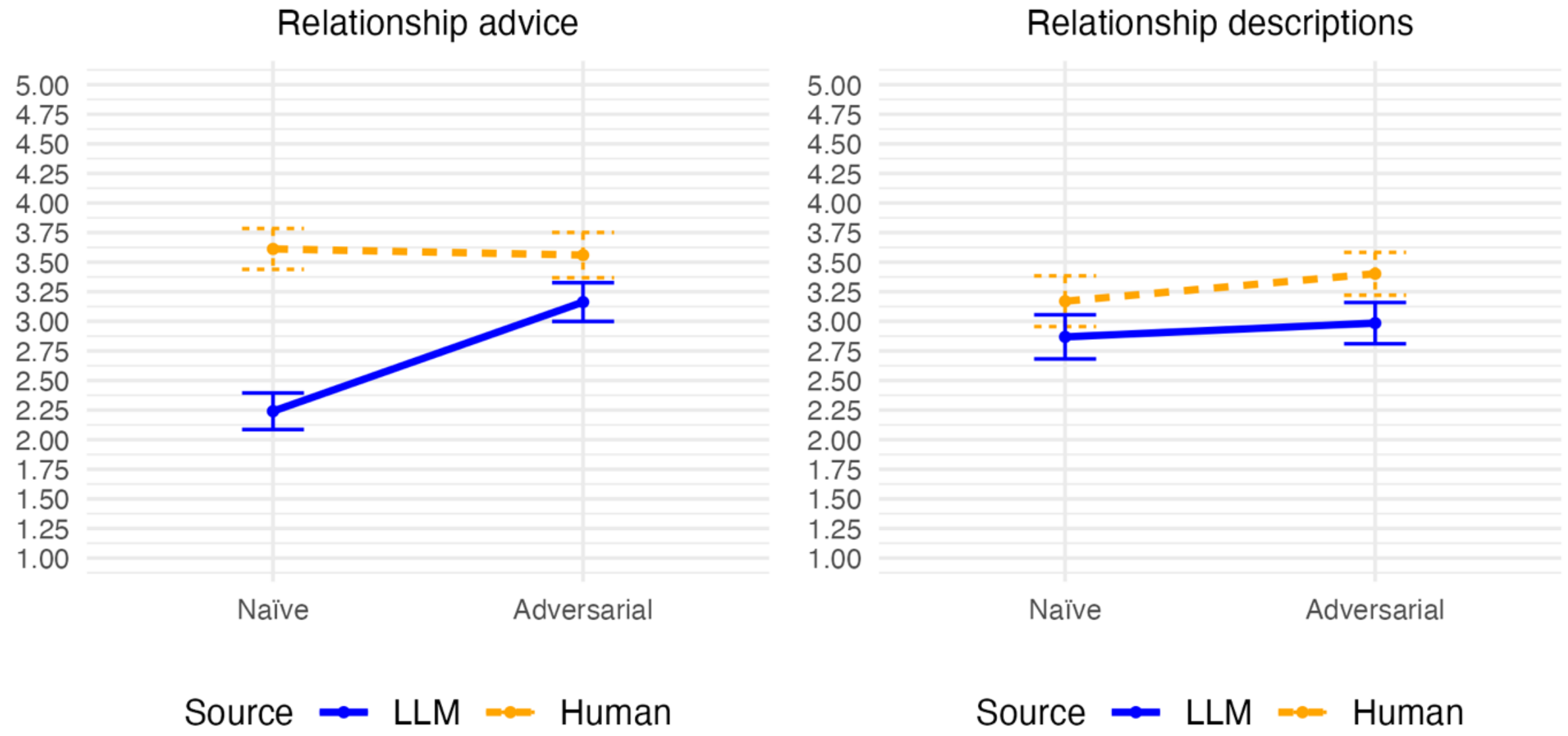

Note. The bars around the means represent the 99% confidence intervals. Source judgment values ranged from 1=definitely AI-generated to 5=definitely human-written. See SM4 for results in table format.

Discussion Study 2

The second experiment adds to evidence that human-written texts are perceived as more human than those written by an LLM, an effect that was twice as pronounced when a task was positioned in a context more relevant for empathy ($d$=1.01) than when it was not ($d$=0.46). The human advantage was thus halved when the task required only describing a relationship rather than providing advice. Contrary to expectations, human participants rated the description task as a little more difficult than the advice task ($d$=0.26) but that difference did not affect the findings (SM7).

We further found that the adversarial instructions were effective only when the task required empathy, and only for the LLM. Here, the second experiment replicated the findings that the LLM (+41.0%), but not humans (-1.4%), responded to the instructions to be as human as possible. The contrast in the effectiveness of the adversarial instructions is striking. The absence of an effect in the adversarial condition for humans suggests that human writers could not increase their humanness perception. It is unclear whether the human writing is evidence of ceiling effects of conveying humanness or attributable to confusing task instructions. Possibly, human participants have not understood how to come across as more human given that they are human. If the task to convey humanness was too hard and confusing, we would expect to find differences between these and contraindicative instructions (e.g., to not come across as an LLM). We, therefore, tested the effect of task instruction phrasing in a preregistered experiment in Study 3 by manipulating whether participants had to convey that they are human or convey that they are not an LLM.

## STUDY 3

The third experiment manipulated the Condition instructions to include the additional phrasing targeted at evading the writing style associated with a language model. We focused solely on relationship advice in a 2 (Source: human vs LLM) by 3 (Condition: adversarial vs naïve vs evasion) between-subjects design. The evasion condition was intended to direct the task towards writing "not like an LLM" and hence removing the – for human writers - potentially confusing instruction to convey humanness. By contrasting that evasion condition with the naïve and adversarial condition, we were able to test two hypotheses.

As a replication of Study 1 and 2, we hypothesised (Hyp. 1) that LLM-generated texts (but not human-written texts) are perceived as more human in the adversarial condition than in the naïve condition. If the adversarial instructions were too conflicting for human writers, we expect that a redirection towards evading an AI model's language would benefit humans but not an LLM. Hyp. 2 stated that for human-written texts (but not LLM-generated texts), the perceived humanness differs between the adversarial and evasion condition (bi-directional).

We added control variables on human-writers empathy and asked in free-text responses for the strategies employed to write relationship advice under the respective condition. A priori sample size calculation - for Cohen's $f$=0.23 (from Study 2), $\alpha$=0.01, power=0.90 – resulted in a total required sample size of 287 for human participants. To account for drop-out and participant exclusion, we aimed to collect data from $n$=450 participants (150 in each condition). After applying data exclusion criteria, we collected the same number of observations from an LLM. We also explored whether the sophistication of the LLM affected the ability to mimic human writing. The preregistration of the design, methods, hypotheses and confirmatory analyses is available at: https://aspredicted.org/2t47-kwzz.pdf

Method

The experimental procedure was identical to Study 2, with the following differences: *i)* we added an evasion condition, *ii)* human participants provided additional control variables, *iii)* we focused solely on relationship advice, and *iv)* human participants answered in a free-text field how they approached the task.

We recruited only participants who had not taken part in any of the data collection in Study 1-2. Data were collected via Prolific from $n$=451 participants (50.33% female, $M_{age}$=38.42, $SD_{age}$=13.07, remuneration: GBP 2.25), who were randomly allocated to the naïve ($n$=146), adversarial ($n$=155) or evasion condition ($n$=150). After excluding those that failed the manipulation check (i.e., recalling their task instructions, $n$=91), we retained $n$=360 participants for the analysis and collected the same number of LLM-generated pieces of relationship advice.

*Evasive vs. adversarial vs. naïve task instructions*

The instructions under the naïve (i.e., no specific instructions) and the adversarial condition (i.e., writing relationship advice that humans will perceive as human-written) remained unchanged and each participant received basic information about a fictitious relationship for which they had to write advice to a friend. In the new evasion condition, participants were instructed to "[...] *write a brief piece of advice to a friend who is having relationship problems. [...] Your goal is to write a text that human participants will not perceive as originating from a large language AI model.*" Humans and the LLM received the same instructions in each condition.

*Refined control questions*

We added two questions about different aspects of empathy required in the task (compassion and perspective taking, rated on a 7-point scale from "completely disagree" to "completely agree") and asked each participant to fill in the *Questionnaire of Cognitive and Affective Empathy* (QCAE, [31]). The QCAE consists of 31 items with affective empathy (12 items) and cognitive empathy (19 items) as subscales. Each item (e.g., “I can easily work out what another person might want to talk about”) is judged on a 4-point scale (1=strongly disagree to 4=strongly agree). We calculated the subscale scores and use these as control variables.

*Free text information about strategies*

After writing their relationship advice, the participants wrote in an open text box which strategy they used to write their relationship advice. We examined these open-ended responses with a topic modelling technique and test whether specific strategies were associated with the conditions and with the humanness ratings.

*Comparing different LLMs*

Since one single LLM is always a snapshot of the capabilities of the models up to that point, we collected data from the GPT-4 model that was used in Study 1 and 2 (gpt-4-0125-preview) as well as from a more advanced model that has been published since then (GPT-4o).

*Source evaluation*

A new sample of *n*=350 participants (48.57% female, $M_{\text{age}}$=40.30, $SD_{\text{age}}$=13.54, remuneration: GBP 2.00) was recruited from Prolific to judge 1080 relationship advice texts (360 texts each from humans and the two LLMs). Participants rated a random selection of ten texts and used the same 5-point scale as in Study 1 and 2 ("1=definitely AI-generated" to "5=definitely human-written"). Each text was, on average, assessed by 3.12 humans (*SD*=0.47).

Results

*Preliminary analysis*

There were no significant differences between the three conditions in any of the control variables, *p*>.01 (SM5). These findings imply that, contrary to what motivated Hyp. 1, we found no significant difference in difficulty between the conditions, suggesting that humans did not find the adversarial instructions more difficult than the evasion or the naïve instructions.

*Human judgments*

The 2 (Source) by 3 (Condition) ANOVA indicated a significant main effect of Source, $F(1, 714)=205.79$, $p<.001$, $\eta^2=0.22$ [0.16; 1.00]. Human-written texts (*M*=3.54, *SD*=0.88) were perceived as significantly more human than LLM-generated texts (*M*=2.67, *SD*=0.81), *d*=1.02 [0.82; 1.22], $BF_{10}$>100, $BF_{01}$<0.01, irrespective of the Condition manipulation (Figure 3).

Further, a significant Condition main effect, F(2, 714)=16.75, p<.001, $\eta^2$=0.04 [0.02; 1.00] revealed that, irrespective of Source, relationship advice written under the adversarial (*M*=3.24, *SD*=0.87) and evasion (*M*=3.24, *SD*=0.86) condition were judged as significantly more human than those written in the naïve condition (*M*=2.88, *SD*=1.04), *d*=0.38 [0.14; 0.61], $BF_{10}$>100, $BF_{01}$<0.01, and *d*=0.37 [0.14; 0.61], $BF_{10}$>100, $BF_{01}$<0.01, respectively. There was no significant difference between the adversarial and evasion conditions, *d*=0.00 [-0.24; 0.24], $BF_{10}$=0.10, $BF_{01}$=9.64.

The significant Source by Condition interaction, $F(2, 714)=20.50$, $p<.001$, $\eta^2=0.05$ [0.02; 1.00], supported Hyp. 1. The LLM-generated texts were judged as significantly more human under the adversarial instructions than under the naïve instructions, $d=1.06$ [0.72; 1.41], $BF_{10}>100$, $BF_{01}<0.01$, while there was no significant difference for human-written texts, $d=0.06$ [-0.27; 0.38], $BF_{10}=0.15$, $BF_{01}=6.53$. Hyp. 2 predicted that human-written texts – but not LLM-generated ones - would respond to the evasion condition. We found no support for that hypothesis since there was no significant difference in humanness perception between the evasion and naïve condition for human-written texts, $d=0.03$ [-0.31; 0.36], $BF_{10}=0.14$, $BF_{01}=6.92$. Contrary to the expectation, LLM-generated texts responded to the evasion condition so that these texts were perceived as significantly more human than those written in the naïve condition, $d=1.05$ [0.70; 1.41], $BF_{10}>100$, $BF_{01}<0.01$. There was no significant difference between the adversarial and evasion condition, for neither humans, $d=0.03$ [-0.31; 0.36], $BF_{10}=0.15$, $BF_{01}=6.76$, nor LLM-generated texts, $d=0.04$ [-0.30; 0.38], $BF_{10}=0.15$, $BF_{01}=6.67$.

These findings were unaffected by the inclusion of the covariates self-reported motivation, difficulty, required empathy, compassion, perspective-taking and QCAE subscales of affective and cognitive empathy (SM7). There was no difference between the two LLM versions (GPT-4 and GPT-4o) tested here (SM8).

Figure 3. *Human judgments for relationship advice by Source (LLM vs human) and Condition (naïve vs adversarial vs evasion) in Study 3.*

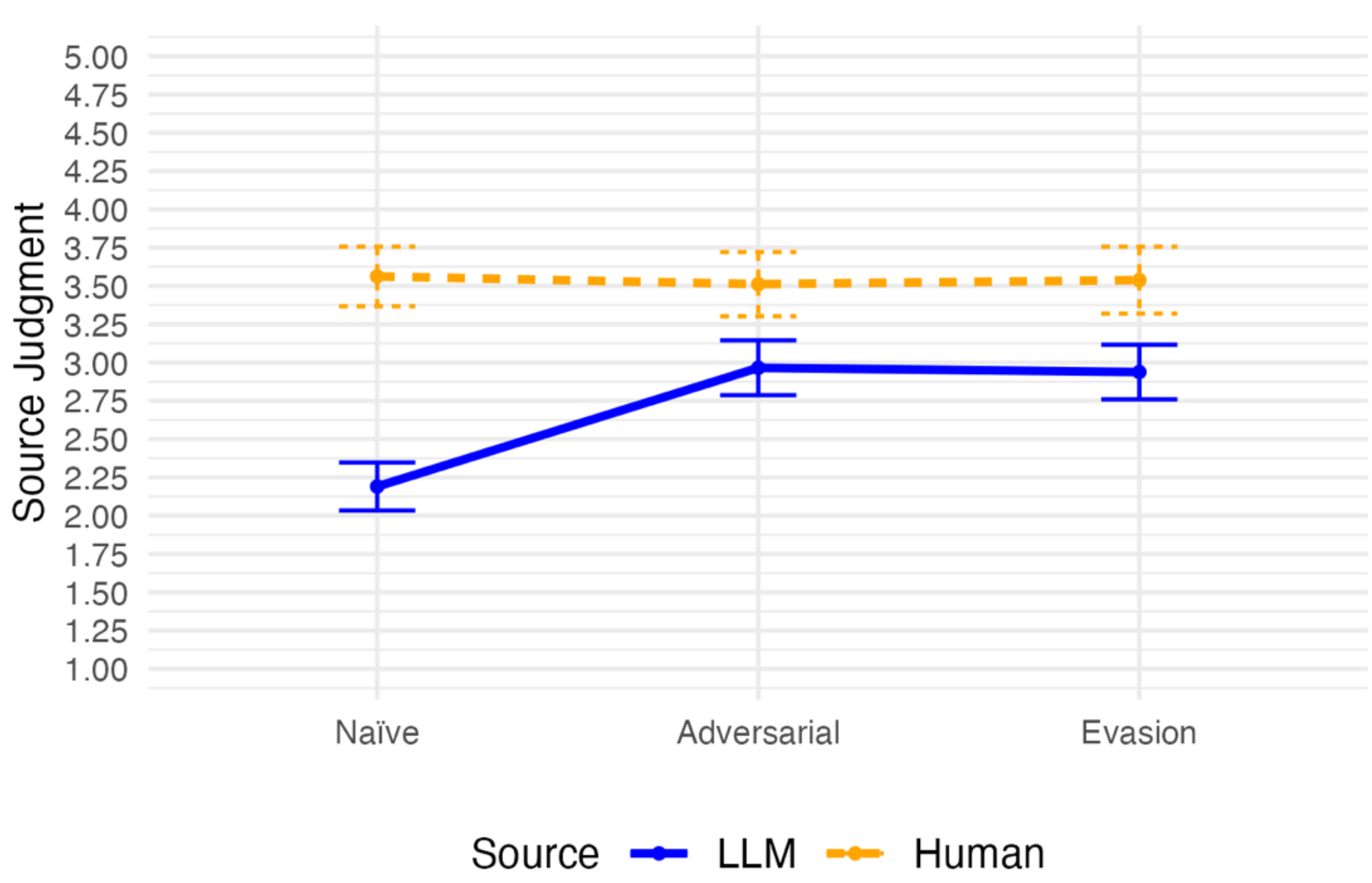

Note. The bars around the means represent the 99% confidence intervals. Source judgment values ranged from 1=definitely AI-generated to 5=definitely human-written. See SM6 for results in table format.

*Diagnostic value of human judgments*

The diagnosticity analysis indicated that in the naïve condition, human judgments were diagnostic in telling human-written from LLM-generated texts, AUC=0.88 [0.83; 0.94] but this dropped significantly in the adversarial, AUC=0.69 [0.60; 0.78], and evasion condition, AUC=0.70 [0.61; 0.79].[3]

*Strategies employed by human writers*

We used an embeddings-based topic modelling approach to identify overarching themes in the free-text responses of human participants when asked about their strategies to write their advice (see SM9 for full details). That approach revealed four topics that we labelled based on the distinctive *n*-grams and the core documents (i.e., free text responses) for each topic. The most common strategy was "perspective taking" (60.00% of the sample), followed by "making it personal" (13.61%), "resorting to one's own experience" (11.94%) and "mentions of cheating" (11.39%). The remaining 3.05% of the responses were assigned to be outliers since they did not fit any topic cluster. There was no association between the use of these topics and the Condition phrasing, $X^2(8)=12.05$, p=.149, so humans did not use specific strategies for the condition requirements.

Discussion Study 3

We tested whether the effects reported in Study 1-2 were due to task instructions that humans found too confusing. The findings of Study 3 suggest that there was no difference between contraindicative (i.e., evading a writing style of a large language model) and adversarial instructions (i.e., conveying humanness). In both conditions, the LLM responded to the instructions and increased the humanness perception of generated relationship advice (adversarial: +36%; evasion: +34%) compared to naïve instructions. Humans showed no such effect (adversarial: -1%; evasion: +0%). These findings supported our replication hypothesis. We found no support for the hypothesis that humans would respond to the evasion instructions.

Taken together, the previous three studies indicate that LLMs show a clear adjustment to task instructions and convey increasing humanness, contrary to humans whose writing remained unaffected by instructions. Importantly, Studies 1-3 assumed that empathy mattered for the display of humanness. When comparing a presumed low-empathy task (i.e., writing relationship descriptions) to a high-empathy task (i.e., writing relationship advice), post-hoc judgments of human authors suggest that required empathy was indeed higher for the high-empathy task. To understand whether empathy is the vehicle through which LLMs increase humanness, however, we need to isolate the effect of empathic language. In Study 4, we manipulated the use of empathy in LLM relationship advice by either explicitly instructing the model to resort to empathy (usage condition) or not including further instructions (neutral condition).

**STUDY 4**

We followed the general procedure of Studies 1-3 with the exception that we only collected LLM responses since the adjustment to adversarial instructions was only observed in LLMs but not in humans. We used the same relationship advice task and human source judgement procedure.

[3] AUC comparison tests: naïve vs adversarial: *D*(394.72)=4.83, *p*<.001; naïve vs evasion: *D*(359.57)=4.52, *p*<.001; adversarial vs evasion: *D*(453.55)=0.17, *p*=.866.

In addition, we also asked human participants to evaluate each text on three empathy dimensions. The experimental conditions were the same as in the previous studies in the naïve and adversarial conditions but were extended by a second experimental factor: empathy use. LLMs were instructed to resort to empathy or not.

The experiment's design, hypothesis and analysis were preregistered (see: https://aspredicted.org/ik8dn7.pdf). We hypothesised that (H1 – replication of Studies 1-3) the perceived humanness score under adversarial instructions is higher than the perceived humanness score under naïve instructions. Further, we expected that if empathy was the driving mechanism for humanness, that (H2) the perceived humanness score under empathy-usage instructions would be higher than the perceived humanness score under neutral empathy instructions. The effects of condition instructions and empathy usage would then also have been expected to interact so that (H3) the difference in perceived humanness score between naïve and adversarial instructions (adversarial > naïve) would be larger when empathy use is required than under neutral instructions.

A priori power analysis for the interaction effect of Condition and Empathy (alpha=0.01, power=0.90, $f$=0.15), suggested $n$=665 required LLM responses.

Method

We generated 672 responses from the GPT-4o model under the 2 (Condition: adversarial vs naive) by 2 (Empathy: usage vs neutral) design. For each combination of the two factors, we collected 168 pieces of relationship advice. Each text was evaluated regarding its source and the degree to which they convey empathy.

*Empathy usage manipulation*

Half of the responses were generated under the naïve and half under the adversarial task instructions – identical to Studies 1-3. In addition, we manipulated the usage of empathy. Half of the prompts included an "empathy usage" instruction, following [17]. These prompts outlined three dimensions of empathy and contained the instruction to use them in the relationship advice: "*[...] Your task is further to make use of empathy in your writing. Empathic relationship advice [...] must include three elements. These are cognitive empathy, which involves understanding how the other person is feeling, affective empathy, which involves sharing and experiencing the same emotions, and motivational empathy, which involves showing you care and want to help them. Use these elements of empathy as best as you can.*" The other half did not contain such empathy usage instructions (i.e., the neutral empathy condition), thereby replicating the design combinations of the previous studies.

*Source evaluation*

A sample of 219 participants (47.95% female, $M_{age}$=43.24, $SD_{age}$=13.20, remuneration: GBP 2.00) recruited from Prolific evaluated the all texts using the same 5-point scale as in Studies 1-3 ("1=definitely AI-generated" to "5=definitely human-written"). Each participant assessed a random selection of ten different statements; each statement was assessed by 3.14 humans ($SD$=0.49), on average.

*Empathy evaluation*

Another sample of human participants was recruited to measure the perceived empathy. Specifically, *n*=232 participants (45.26% female, $M_{age}$=42.33, $SD_{age}$=12.44, remuneration: GBP 2.25) read and evaluated each piece of relationship advice on cognitive, emotional and motivational empathy (on a 5-point Likert scale ranging from 'strongly disagree' to 'strongly agree') on these items: "The author seems to detect the emotional state of the advisee" (cognitive), "The author seems to experience similar emotions or situations as the advisee" (emotional), "The author seems to be motivated to help the advisee" (motivational). Each participant assessed ten randomly selected pieces of relationship advice on all three dimensions and for each text, an attention check was included asking them to select one of the extreme values of the scale. Data quality was high with only 48 trials (0.57%) being excluded due to failed attention checks at the trial level. Each text was assessed on all three empathy dimensions by 3.11 humans, on average (*SD*=0.85).

Results

*Human judgments*

The 2 (Condition) by 2 (Empathy) ANOVA (Figure 4, SM10) showed significant main effects of Condition, $F(1, 668)=53.22$, $p<.001$, $\eta^2=0.07$ [0.03; 1.00], and Empathy, $F(1, 668)=44.42$, $p<.001$, $\eta^2=0.06$ [0.03; 1.00]. Texts written under the adversarial instructions (*M*=3.04, *SD*=0.81) were perceived as significantly more human than those written under naïve instructions (*M*=2.60, *SD*=0.80), *d*=0.55 [0.34; 0.75], $BF_{10}$>100, $BF_{01}$<0.01. The Empathy main effect indicated that texts written with explicit usage instructions (*M*=2.62, *SD*=0.83) were perceived as less human than texts written under neutral empathy instructions (*M*=3.02, *SD*=0.79), *d*=0.50 [0.29; 0.70], $BF_{10}$>100, $BF_{01}$<0.01. There was no significant interaction between Condition and Empathy, $F(1, 668)=0.02$, $p=.898$, $\eta^2=0.00$ [0.00; 1.00].

Figure 4. *Human source evaluation for relationship advice by Condition (naïve vs adversarial vs evasion) and Empathy (usage vs neutral) in Study 4.*

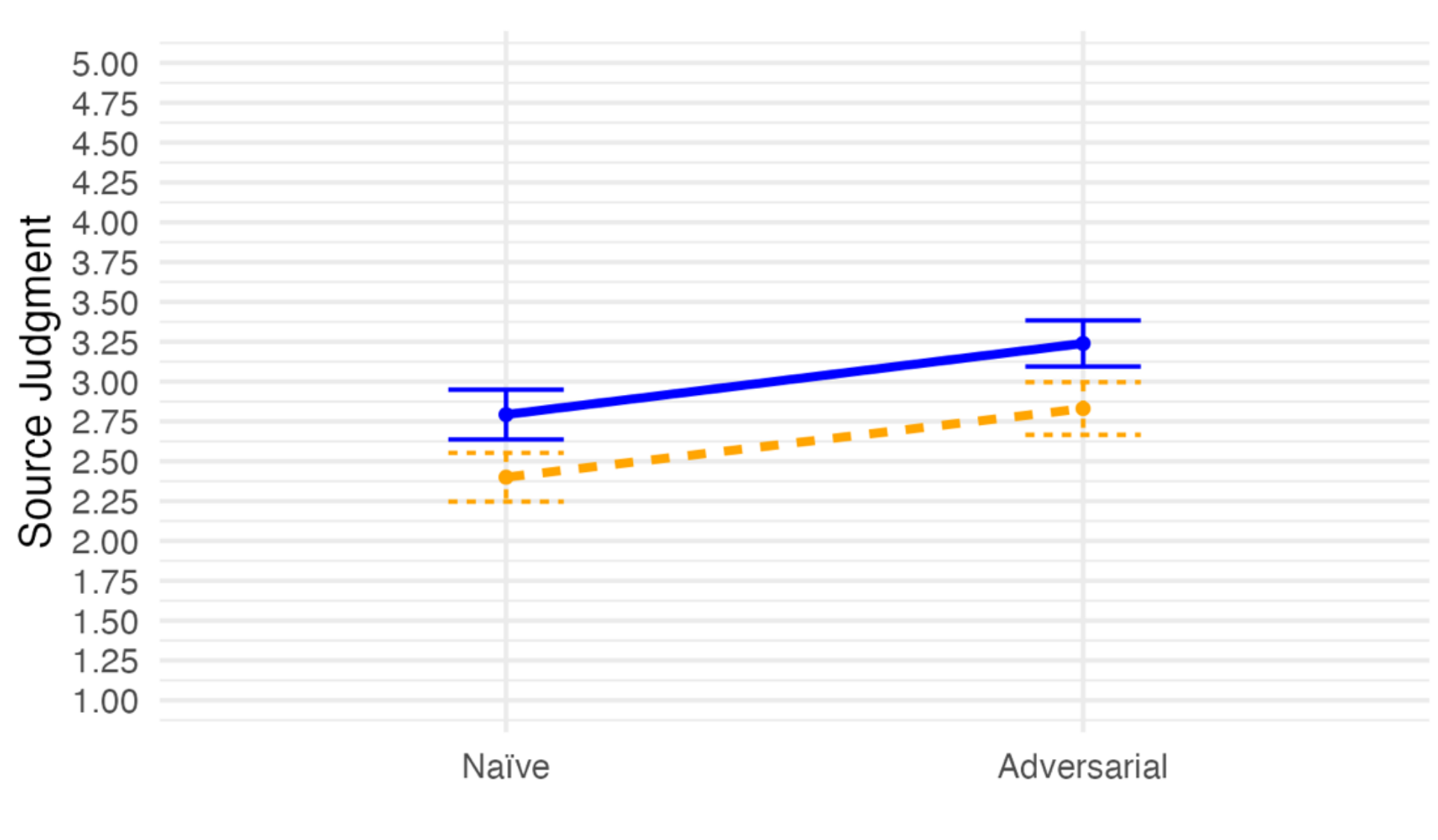

Note. The bars around the means represent the 99% confidence intervals. Source judgment values ranged from 1=definitely AI-generated to 5=definitely human-written. See SM10 for results in table format.

*Empathy evaluation*

We ran the same 2 (Condition) by 2 (Empathy) ANOVA on the perceived empathy scores on each of the three empathy dimensions. For all dimensions of empathy, there was a significant main effect of Empathy[4], which suggested that texts written under the empathy-usage instructions were scored significantly higher on these dimensions than texts written under neutral instructions (Figure 5, SM11), $d$=0.90 [0.60; 1.11], $BF_{10}$>100, $BF_{01}$<0.01 (cognitive empathy), $d$=0.64 [0.43; 0.83], $BF_{10}$>100, $BF_{01}$<0.01 (emotional empathy), $d$=0.29 [0.09; 0.49], $BF_{10}$=90.06, $BF_{01}$=0.01 (motivational empathy).

For emotional and motivational empathy, there was further a significant interaction between Condition and Empathy[5], suggesting that the effect of using empathy was more pronounced in the naïve task instructions (emotional empathy: $d$=0.90 [0.61; 1.20]; motivational empathy: $d$=0.53 [0.24; 0.81]) than under adversarial instructions (emotional empathy: $d$=0.38, [0.00; 0.66]; motivational empathy: $d$=0.07, [-0.21; 0.35]. There was no significant main effect of Condition on any of the empathy scores.

Figure 5. *Human empathy evaluation for relationship advice by Condition (naïve vs adversarial) and Empathy (usage vs neutral) in Study 4*.

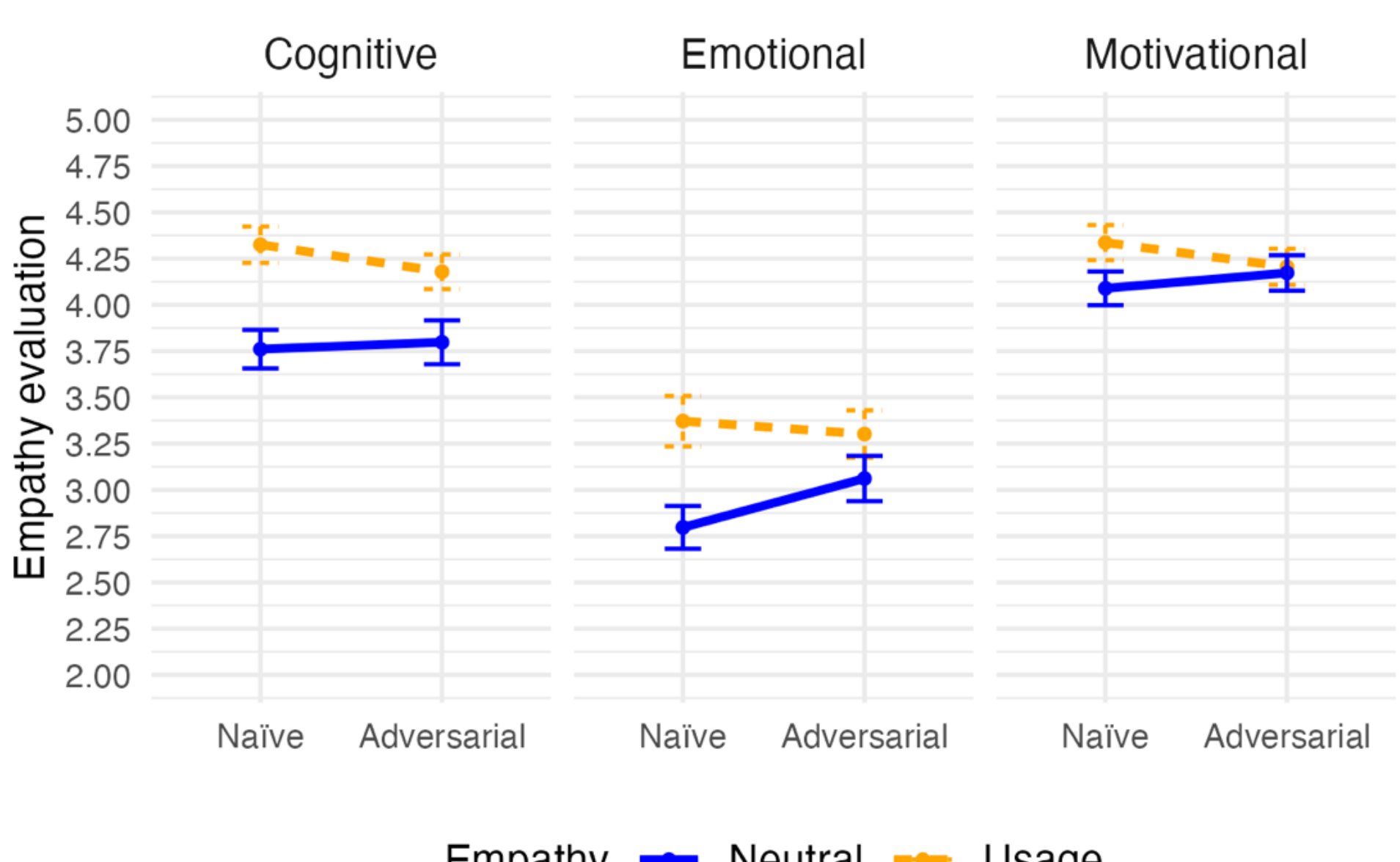

Note. The bars around the means represent the 99% confidence intervals. Empathy evaluations ranged from 1=strongly disagree to 5=strongly agree. See SM6 for results in table format.

[4] Cognitive: $F(1, 668)$=136.20, $p$<.001, $\eta^2$=0.17 [0.13; 1.00]; Emotional: $F(1, 668)$=69.34, $p$<.001, $\eta^2$=0.09 [0.05; 1.00]; Motivational: $F(1, 668)$=14.48, $p$<.001, $\eta^2$=0.02 [0.01; 1.00])

[5] Emotional: $F(1, 668)$=11.66, $p$=.001, $\eta^2$=0.02 [0.00; 1.00]; Motivational: $F(1, 668)$=8.45, $p$=.004, $\eta^2$=0.02 [0.00; 1.00].

Discussion Study 4

We tested whether using empathic language is the mechanism through which humanness is conveyed by LLMs. The findings replicated the key finding from Studies 1-3 that LLMs can effectively increase the humanness perception on demand (+16.92%), supporting Hypothesis 1. We found no evidence for empathy as the driving mechanism for humanness (Hyp. 2 and 3): contrary to our expectation, humanness perception was lower when the instructions specifically required the use of empathy. The absence of the empathy mechanism in conveying humanness cannot be explained with the LLMs inability to simulate empathic writing, since instructions to use empathy effectively increased perceived empathy on all three empathy dimensions. Put differently, the LLM can increase humanness on demand and can craft empathic writing on demand – on all three empathy dimensions. But to convey humanness, the LLM does not resort to empathic writing. This indicates that empathic phrasing itself is not sufficient for producing humanness perceptions. If not through empathy usage, how does the LLM achieve the increase in humanness perception instead? In Study 5, we employ computational text analysis techniques to explore how the LLM changes its language when trying to appear human.

## STUDY 5

We look at the linguistic strategies humans and LLMs use to appear human through linguistic differences. For the subsequent analyses, we combine the data from all four experiments from the naïve and adversarial relationship advice tasks (1085 human-written, 1421 LLM-generated[6], corpus size: 489k words). We assess how humans and LLM changed the relationship advice when instructed to appear as human as possible – which humans failed to do effectively, and LLMs were particularly good at. For these analyses, we look at term frequencies (through an *n*-gram differentiation test) and psycholinguistic differences as measured by the Linguistic Inquiry and Word Count (LIWC) software and an empathic concern dictionary. We also examined the role of spelling mistakes and word frequency as potential cues and examined whether humans resorted to such decision heuristics.

### Method

*N-gram differentiation*

The *n*-gram differentiation test [32] statistically compares frequencies of unigrams, bigrams and trigrams with a non-parametric signed rank sum test approach. Ties in ranks are resolved by random ranks in 500 iterations, which are averaged. For parsimony, only *n*-grams that appeared in at least 5% of all documents were included, stop words were removed, and each term was reduced to its word stem. The effect size for the frequency comparisons is *r* (ranging from -1.00 to 1.00).

*LIWC analysis*

The LIWC (version 2022)[33] measures the proportion of words in a document that belong to a range of predefined linguistic and psycholinguistic categories using a word-matching approach with curated dictionaries. Categories include cognitive processes, emotions, and social processes. We conducted bottom-up testing with all 117 variables using a Bonferroni-corrected significance threshold.

[6] The LLM-generated texts included here stem from the GPT-4 model that was used in Study 1-3. Study 4 included only LLM-generated texts (from the GPT4o model), resulting in the imbalance between human-written and LLM-generated data.

Empathic concern dictionary

We employed the empathic concern dictionary [34] to obtain one single metric that measures the level of empathic concern at the document level using a dictionary of 9,356 terms with empathy ratings.[7]

*Spelling mistakes and vocabulary frequency*

Related work suggests that mistakes and rare words are relevant features between AI and human-generated content [9][8]. We used the raw textual data and counted the number of spelling mistakes using the *hunspell* R package [35]. To avoid inflating the number of mistakes and to correct for statement length, we counted mistakes for both British and American English spelling, took the larger of the two, and standardised the counts to mistakes per 100 words. The vocabulary frequency was obtained by calculating the average rank frequency of the words in a document. As a reference, we used the most frequent 10k words based on Google's Trillion Word Corpus[9]. Lower rank scores imply a higher frequency. For both measures, we examine the diagnostic value and their usage as cues by humans. If humans used a cue in their judgment, we expect high correlations between the cue and the human judgment.

## Results

*Human strategies to appear human*

Differences in human-written relationship advice between the naive and the adversarial condition were expectedly minor (SM12). Merely four *n*-grams emerged as different (all adversarial > naive): "know", "can", "right" and "just". Similarly, the LIWC analysis revealed that only two categories were significantly more prominent in the adversarial than naïve condition (focus on the present, mentions of auditory words). Conversely, the use of third person plural (LIWC category "they") and the use of longer words (more than six characters) were somewhat more pronounced in the naive than in the adversarial condition. The results align with our finding that humans were generally unable to appear more human.

*LLM strategies to appear human*

The substantial effects we observed when LLMs moved from the naive to the adversarial condition (Study 1: $d$=1.24; Study 2: $d$=1.21, Study 3: $d$=1.06, Study 4: $d$=0.55) manifested in textual differences. Table 1 shows that the LLM reduced the "dear friend" greeting and instead switched to a more informal "hey". We also observe the increased use of "really sorry" and fewer formal references to the relationship (SM12).

Furthermore, when trying to appear human, the LLM resorted to considerably more netspeak (e.g., "lol", "haha") and conversational markers (e.g., "yes", "yeah", "okay"). The temporal focus shifted to the present, and the language became more engaging (e.g., an increased use of verbs) and contained more self-references (first person singular). At the same time, the LLM dropped the use of big words and references to drives – particularly power and affiliation as drives – and became less analytic. These findings provide strong evidence that the LLM aligned the content and style to the need to appear more human by shifting to informal language with self-references.

---

[7] The empathic concern dictionary is only available in the LIWC implementation and was not retrievable online elsewhere.

[8] Note that [9] found that mistakes and rare phrasing were misaligned features (i.e., humans associated these with AI-generated content while both features were, in fact, indicative of human writing).

[9] https://github.com/first20hours/google-10000-english

Table 1. *Top 10 features by effect size for the n-gram differentiation and LIWC analysis for LLM-written relationship advice between the adversarial and naive condition*

| *n*-gram differentiation | | | | LIWC | | | |
|---|---|---|---|---|---|---|---|
| Adversarial > naive | | Naive > adversarial | | Adversarial > naive | | Naive > adversarial | |
| *n*-gram | *r* | *n*-gram | *r* | Category | *d* | Category | *d* |
| hey | 0.84 | dear | -0.79 | netspeak | 2.34 | BigWords | -2.47 |
| realli ("really") | 0.59 | dear_friend ("dear friend") | -0.79 | Linguistic | 2.29 | Drives | -1.54 |
| go | 0.46 | friend | -0.64 | function | 2.16 | affiliation | -1.41 |
| tri ("try", "trying") | 0.44 | challeng ("challenging") | -0.54 | Conversation | 1.86 | Analytic | -1.30 |
| tough | 0.44 | dear_friend_face ("dear friend, facing [...]") | -0.43 | verb | 1.84 | tone_pos | -1.06 |
| sorri ("sorry") | 0.40 | friend_face ("friend, facing [...]") | -0.41 | auxverb | 1.76 | Affect | -1.03 |
| thing | 0.40 | approach | -0.40 | focuspresent | 1.51 | power | -0.79 |
| realli_sorri ("really sorry") | 0.40 | reflect | -0.36 | i | 1.31 | article | -0.78 |
| long_time ("long time") | 0.37 | navig ("navigating") | -0.34 | adverb | 1.17 | det | -0.72 |
| hey_first ("hey, first") | 0.46 | communic ("communication", "communicating") | -0.32 | ipron | 1.13 | OtherP | -0.70 |

*Note*. The effect size *r* for the *n*-gram comparison ranges from -1.00 to 1.00. The Cohen's *d* effect size represents small ($d$=0.20), moderate ($d$=0.50) and large effects ($d$=0.80)[36,37].

*Empathic concern*

There was only a significant main effect of Source, $F(1, 2502)=1067.44$, $p<.001$, $\eta^2=0.30$ [0.27; 1.00], indicating that human-written relationship advice (*M*=91.44, *SD*=2.78) contained a higher level of empathic concern than LLM-generated advice (*M*=88.58, *SD*=1.54), *d*=1.32 [1.20; 1.43]. There was no significant effect of Condition or interaction. Irrespective of the task instructions, humans' language was more empathic than that of LLMs.

*Spelling mistakes and vocabulary frequency*

For spelling mistakes, there was merely a significant main effect of Source, $F(1, 2502)=364.31$, $p<.001$, $\eta^2=0.13$ [0.10; 1.00] indicating that human-written texts contained a higher rate of spelling mistakes (per 100 words, *M*=0.77, *SD*=1.25) than LLM-generated ones (*M*=0.12, *SD*=0.26; *d*=0.77 [0.66; 0.88], Table 2, SM10 for full results). That effect translated to moderate diagnostic values (AUCs between 0.69 and 0.73). Small non-parametric correlations between the rate of mistakes and the human source evaluation score suggest that humans did not use spelling mistakes as a cue in their judgment.

In contrast, the mean rank vocabulary frequency was highly diagnostic, with close to perfect diagnostic power (AUCs between 0.92 and 0.96). There was a significant main effect of Source, $F(1, 2502)=3053.50$, $p<.001$, $\eta^2=0.55$ [0.52; 1.00] showing that humans used a less complex vocabulary (*M*=706.61, *SD*=193.39) than the LLM (*M*=1116.12, *SD*=202.91), *d*=2.06 [1.93; 2.19].

The significant main effect of Condition, $F(1, 2502)=287.10$, $p<.001$, $\eta^2=0.10$ [0.08; 1.00] indicated a more complex vocabulary in the naïve ($M=1000.82$, $SD=314.32$) than in the adversarial condition ($M=876.52$, $SD=234.40$), $d=0.45$ [0.34; 0.55]. A significant interaction between Source and Condition subsumed these main effects of Condition and Source (SM12), $F(1, 2502)=141.10$, $p<.001$, $\eta^2=0.05$ [0.03; 1.00], and revealed the LLM's strategy to resort to lower complexity language, $d=1.14$ [0.99; 1.28], and absence thereof by humans, $d=0.13$ [-0.03; 0.28].

Interestingly, there was substantial negative correlation ($\rho=-0.56$, $p<.001$) between human judgments (higher scores being indicative of "more humanness") and the rank frequency in the naïve condition, suggesting that humans judged relationship advice as "more human" when the frequency rank was lower (i.e., less complex language was used). That cue use was only half as a pronounced in the adversarial condition ($\rho=-0.27$, $p<.001$).

Table 2. *Descriptive statistics, diagnostic value and cue use of spelling mistakes and mean vocabulary frequency per condition in Study 5.*

| | | Spelling mistakes per 100 words | | | Vocabulary rank frequency | | |
|---|---|---|---|---|---|---|---|
| Source | Condition | *M* (*SD*) | AUC | *ρ* | *M* (*SD*) | AUC | *ρ* |
| Human | Naive | 0.81 (1.41) | 0.69 [0.65; 0.72] | 0.19* | 718.88 (205.97) | 0.96 [0.94; 0.97] | -0.56* |
| GPT4 | Naive | 0.15 (0.29) | | | 1216.23 (186.29) | | |
| Human | Adv. | 0.72 (1.06) | 0.73 [0.70; 0.76] | 0.07* | 694.27 (179.20) | 0.92 [0.90; 0.94] | -0.27* |
| GPT4 | Adv. | 0.09 (0.23) | | | 1015.59 (165.91) | | |

*Note.* *=significant Spearman's $\rho$ correlation at $p<.01$. Spearman's correlations are between human judgements of source and spelling mistakes / vocabulary complexity.

Discussion Study 5

Our computational text analysis uncovered the traces that humans and LLM left when they were instructed to appear as human as possible. Humans generally failed said task, reflected by barely any linguistic differences. In contrast, the LLM aligned generated output to the task requirement: the relationship advice became more informal, less analytic, and the chosen vocabulary became more common and less complex. However, the LLM did not seem to increase humanness through utilising empathic language but, instead, relied on surface-level stylistic mimicry of humanness. Merely the number of spelling mistakes was not aligned with human writing, so LLMs kept producing near mistake-free text. Put differently, the LLM seemed to have a good representation of how it could come across as more human and succeeded at changing human perception.

## GENERAL DISCUSSION

Large language models (LLMs) have rapidly been endorsed in academic research. Some studies report that LLMs generate content nearly indistinguishable from humans' content [8,9]. Comparisons between human and AI models typically involve text sourced from humans in a context where they were unaware of an AI's masquerading abilities and had little incentive to appear particularly human (e.g., host profiles) and may thus be unfair to human language ability. Furthermore, we argued that the tasks on which LLMs and humans were compared commonly lack the requirement for empathy. The current paper sought to correct this by experimentally disentangling how motivation to appear human and using empathy affected the perception of human and LLM-written content.

*Humans being humans*

In three experiments (Studies 1-3), humans produced text that was perceived as more human than texts generated by an LLM under identical instructions. When the writing task required empathy, humans were superior in expressing their humanness ($d$=0.82, $d$=1.01 and $d$=1.02 in Study 1-3, respectively; all Bayes Factors $BF_{10}$ > 100 indicating extreme evidence in favour of humans) and were perceived as such by other humans. These findings contradict claims about the indistinguishability between human and LLM-generated content [8,9]. Our work also identifies an important role for empathy. In other words, when we require humans to use empathy, their ability shines through, and LLMs struggle to keep up. Under such conditions, third persons can confidently differentiate between generated and original human text.

The condition most similar to related work [9], where LLMs and humans were reported to be practically indistinguishable, required our participants to write a relationship description - rather than providing advice - without any specific instruction or incentive to appear human. Under these conditions, we also find an advantage for the human but to a lesser extent (writing a description: $d$=0.46 vs writing advice: $d$=1.10) with the diagnostic value of third-person judgments bordering the chance level (AUC: 0.60 [0.50; 0.70]). This adds evidence for the role of empathy: when none is required, LLMs indeed seem to produce text close to indistinguishable from humans.

*LLMs pretending to be humans*

Aside from empathy, awareness of the need to appear human could further help understand how humans express their humanness. Our findings revealed unexpected dynamics when we instructed humans and LLMs to appear as human as possible. Contrary to expectations, this adversarial instruction did not widen the gap between humans and LLMs in favour of humans. Instead, only the LLM benefitted from it and was able to adjust and align the generated content so that it approached human-written text. That effect persisted even when we instructed humans and LLM to avoid the writing style of an LLM. The key effect replicated in all three experiments and was substantial: humanness judgments increased by 17-41% for LLMs but remained unchanged for texts written by humans. While the human-written texts were nevertheless still judged as somewhat more human, that gap has drastically narrowed with the diagnostic ability to differentiate between human and LLM dropping markedly (Study 1: drop from AUC=0.83 to AUC=0.62; Study 2: drop from AUC=0.88 to AUC=0.62; Study 3: drop from AUC=0.88 to AUC=0.69). The LLM became better at emulating humanness, both when encouraged to portray humanness and in a scenario where we instructed it to avoid sounding like an LLM.

The other important finding of our work is that humans were unable to increase their humanness perception. The candidate explanation that conveying humanness is paradoxical for humans was tested and ruled out in the third experiment. If a focus on being particularly human was too confusing, we would expect that a different phrasing focused on avoiding writing as an LLM results in a somewhat increased humanness perception. The findings suggested that this was not the case: neither were the adversarial and evasion instructions different in difficulty, nor did they result in different humanness ratings. Another potential explanation for why humans failed to convey more humanness could be a ceiling effect implying that humanness was already exhausted without specific instructions, rendering even prompts towards more humanness as well as those against an LLM writing style ineffective. While for a traditional ceiling effect, we would have expected human-written texts to be rated at the higher end of the 5-point scale on source judgment, the experiments in Study 1-3 suggest that the theoretical maximum might not be the same as a practical maximum of humanness judgments. Expecting genuine human texts to be at the top of the humanness scale might create false expectations.

The striking finding that LLMs are highly responsive to humanness instructions led us to explore how the attempt to pretend to be human manifested in linguistic traces.

*Linguistic traces of humanness*

There is evidence that humans use a mix of aligned and misaligned cues to decide whether a text is LLM-generated [9,38]. For example, in related work, humans correctly associated conversational words with humans but incorrectly expected rare or long word usage from LLMs in previous research [9]. Our work provides conflicting evidence: under the condition where texts were easiest to identify (relationship advice under naive instructions), generated text was aplenty with long words, somewhat rare phrasing ("dear friend") and generally contained less frequent vocabulary (i.e., rarer words), which third-person assessors may have picked up on. Spelling mistakes also emerged as a cue, with humans making more mistakes than LLMs (see also [9]). However, human judges did not seem to rely on that cue and failed to exploit its diagnostic value. The starkest cue was the use of rare terms (i.e., a more complex vocabulary) by LLMs compared to humans, and our results indicate that humans used this heuristic to some extent but not decisively. Had they been able to perceive and use that cue, humans would have been able to tell LLM from human-written text with near-perfect accuracy.

Using term frequency and psycholinguistic analysis offered a view of how our work's adversarial dynamics unfolded. As expected from failure to move the perception score, humans showed no noteworthy linguistic traces when trying to be more human. When LLMs pretended to be human, the strategies revealed a remarkable awareness of what makes a text human, so much so that human judgments increased considerably. The LLM shifted to more conversational language using an informal tone, self-referencing and focusing on the present. The vocabulary was altogether adjusted to use more common words. These changes came at the expense of using long words, appeals to power and affiliation as drives and rather stiff greetings when starting the relationship advice text. These findings align with previous work [9], where human readers used conversational language, self-representations with references to prior experience and a warm tone - rather than monotonous phrasing - to infer humanness. Others have referred to these cues as heuristics humans use to tell generated from original content.

Our work indicates that an LLM pretending to be human may have an implicit representation of these heuristics that allow it to convey humanness. Drawing on research that examined linguistic properties of empathy and compassion in human social media language [14] allows us to speculate about the nature of the shift towards humanness in LLMs. That work disentangled empathy from compassion and identified psycholinguistic correlates of "empathy without compassion" and "compassion without empathy". Our findings show surprising correspondence: under the adversarial instructions, the LLM increased the use of those categories associated with "empathy without compassion" (e.g., first person singular, focus on the present, verbs) and significantly dropped the use of those associated with the "compassion without empathy" (e.g., affiliation, drives, positive tone) [14]. At least from the psycholinguistic analysis, the LLMs' strategy to convey humanness seems to align with faking empathy without compassion. Interestingly, [14] measured empathy with the Empathy Index [39] which focuses on emotional empathy ("feeling what others feel"). Considering these findings, the LLM behaviour we observe can best be described as effectively faking emotional empathy at the expense of showing concern for the other. The exact nature of that linguistic pattern in LLMs should be a topic of future work.

*Stochastic empathy*

Our findings paint a picture of LLMs approaching empathic writing abilities comparable to humans. While some cautioned against anthropomorphising [19], others argue that concepts traditionally reserved for psychological science enter the realm of AI research [20]. The current work shows an LLM's chameleonic abilities: it can mimic a writing style associated with a particular humanness and does so particularly well in tasks that require empathy.

While this is inevitably a learned statistical representation of empathic writing – that is, a stochastic parrot [40] producing seemingly human language without any understanding of what the nature of empathy or the effect of the words generated is [41] – it is noteworthy that the LLM seems to fall back to an implicit representation of humanness. With a simple experimental manipulation, we invoked that humanness in LLMs and found no comparable adaptation to that same manipulation in humans.

The driver of that humanness is not necessarily empathy. Our fourth study showed that an LLM can increase empathy on demand, yet this seemed orthogonal to perceived humanness. While the model is capable of inserting empathic language, such simulated empathy does not increase humanness perception. We argue that this is the key finding of our work: while the LLM was merely generating stochastic empathy, it does seem to hold statistical representations of very *human* language that it used to great effect in conveying humanness. Stochastic empathy enables the model to reproduce linguistic surface markers of empathy without access to the interpersonal consequences of empathic communication and is thus evident linguistically but not functional. Put differently, stochastic empathy produces empathy without humanness and humanness without empathy.

Whereas Studies 1-3 suggested that empathy might underlie LLM humanness, the experimental manipulation of empathy shows that the ability to generate empathic language is neither necessary nor sufficient for increasing humanness perception. The dissociation that empathic language is used statistically without an understanding of empathy is the defining characteristic of stochastic empathy.

The consequences of stochastic empathy are as diverse as they are far-reaching. Some argued that an AI system can only ever emulate cognitive empathy without emotional empathy and would thus aptly be called "psychopathic" [13]. Our findings suggest that, in written communication such as relationship advice, faking a dysfunctional stochastic empathy on all empathy dimensions is no longer out of reach for AI systems. Others point to the double-edge sword of indefatigable empathic AI [15] that offers immediate, low-cost and constant displays of empathy without some of the negative sides of human empathy [12]. But an indefatigable empathic AI also undermines genuine human empathy because when recipients of empathy become accustomed to omnipresent empathy for their concerns, they might hold fellow humans to unrealistic expectations, which could cause tensions in areas such as medical care [11,12]. Indeed, recent evidence suggests that humans' interactions with AI systems show negative spillover to interactions with other humans who are resultingly judged more harshly and treated more demandingly [42]. More recent work further indicates that humans tend to perceive AI-generated responses to recognise their emotions (i.e., "feeling heard") more than human strangers do [16]. However, that effect is quickly compensated by an aversive attitude towards AI responses to emotionally complex situations (see also [17]).

Another implication of LLMs' stochastic empathy is that with an increasing blurriness between AI-generated and genuine human language, the risks for humans to be purposefully misled also increase. Consequences include efforts to defraud individuals [43,44] and modify deceptive statements to appear truthful to humans [45]. While our findings show that humans are still perceived as humans in writing, we also see how that gap between humans and AI systems narrows when we instruct an LLM to deceive. If automated messages can be guided to imitate that humanness, trust in online communication might decrease when humans are wary of generated content. In addition to some of the positive consequences (e.g., low-cost, immediate and constant empathy), the negative impact of stochastic empathy is non-negligible as it could affect the social fabric by increasing distrust through diluting genuine humanness in written communication and by projecting expectations of indefatigable AI empathy onto other humans. Evoking humanness in AI models can thus lead to desirable and helpful consequences but may also create new challenges in the interaction with digital content.

*Limitations and outlook*

Several limitations of the current work are relevant for future work. First, one can argue that the task for the LLM was too difficult because we never exposed the LLM to any human examples. Others have fine-tuned the LLM (i.e., exposed it to known human content for a specific task) and report that this does move the dial towards humanness even further [9]. While such work would likely add to the body of work on LLMs' capabilities, it would also distort inferences about *a priori* representations within the models and add little to an understanding of how an LLM operates. A possible avenue would instead be to learn more about mechanisms that enable empathic writing or even localise these as features in a language model's network [46,47]. In addition, it would be worthwhile to systematically assess how LLMs achieve human-likeness with increasing model sophistication. We did not find a difference when we used a more advanced LLM but evidence on "emergent LLM abilities" [48,49] suggests that increasing model size and sophistication might play a role when examined more comprehensively. Future work could assess humanness of LLM-generated text as a function of model sophistication.

Second, there is also the possibility that the task was too difficult for humans. Even though we offered an instruction (Study 3) that targeted LLM-evasion, humans might simply not have been knowledgeable enough to know exactly what kind of writing style is associated with an LLM. Previous work suggests that humans simultaneously hold correct and incorrect beliefs about linguistic cues associated with AI models [9] and more recent work shows that humans think of AI systems in metaphors as diverse as seeing AI as a teacher, friend, child or search engine [50]. Future work could put the "human humanness limit" to a stricter test and compare groups who either received detailed instructions about what others associate with humanness and LLM-generated content or not. Failure to move humanness perception even with training and detailed instructions would further corroborate the idea of humanness limit in language.
Finally, these findings are limited to the English language and the model's ability to effectively use English. Despite the multilinguistic abilities shown [51], the nuance required in personal, empathic writing may be harder to mimic in other languages. For example, the distinction between personal and impersonal second-person pronouns can be glanced over in English (i.e., "you") but is essential in German, Dutch and Italian. Subtle misalignments (e.g., using "Sie" – the German impersonal, formal "you") in relationship advice to a friend would be a telling sign of no human writing. Future work could extend ours to various languages.

**Conclusion**

Large language models reportedly generate text indistinguishable from humans. We borrowed experimental research approaches from psychology to test whether this finding tolerates conditions favourable to human language understanding. Contrary to our expectation, language models, but not humans, adjusted their writing noticeably when instructed to be particularly human. When humanness was required, the language model seemed to resort to existing representations of empathic human language as informal, simple, self-referencing and focused on the present.